\def\year{2022}\relax
\documentclass[letterpaper]{article} 
\usepackage{aaai22}  
\usepackage{times}  
\usepackage{helvet}  
\usepackage{courier}  
\usepackage[hyphens]{url}  
\usepackage{graphicx} 
\usepackage{natbib}  
\usepackage{caption} 
\DeclareCaptionStyle{ruled}{labelfont=normalfont,labelsep=colon,strut=off} 
\usepackage{booktabs}
\usepackage{bibentry}
\usepackage{xcolor}

\usepackage{newfloat}
\usepackage{listings}

\usepackage{xspace}
\usepackage{algorithm2e}
\usepackage{booktabs}
\usepackage{amsmath,amsfonts,bm}
\usepackage{multirow}
\usepackage{soul}
\definecolor{Red}{rgb}{1,0,0}
\definecolor{Green}{rgb}{0,1,0}
\definecolor{Blue}{rgb}{0,0,1}
\definecolor{Red}{rgb}{0.9,0,0}
\definecolor{Orange}{rgb}{1,0.5,0}
\definecolor{yellow}{rgb}{0.65,0.6,0}
\definecolor{cadmiumgreen}{rgb}{0.2, 0.7, 0.24}

\newcommand{\eat}[1]{}

\newcommand{\V}[1]{\mathbf{#1}}

\newcommand{\X}[1]{\mathbf{\mathrm{S}}}
\newcommand{\Z}[1]{\mathbf{\mathrm{C^-}}}
\newcommand{\VV}[1]{\mathbf{\mathrm{C^+}}}
\newcommand{\W}[1]{\mathbf{\mathrm{M^-}}}
\newcommand{\U}[1]{\mathbf{\mathrm{S^-}}}
\newcommand{\Y}[1]{\mathbf{\mathrm{M^+}}}
\newcommand{\LL}[1]{\mathbf{\mathrm{H^-}}}
\newcommand{\M}[1]{\mathbf{\mathrm{H^+}}}

\newcommand{\proscriptgen}{\textsc{proscript}$_{gen}$\xspace}

\newcommand{\eg}{e.g.,\xspace}

\newcommand{\badg}{y_{bad}}
\newcommand{\corrg}{y_{good}}

\newcommand{\fb}{fb}
\newcommand{\sample}{$(\badg, \fb \rightarrow \corrg)$\xspace}
\newcommand{\badgm}{$y_{bad}$\xspace}
\newcommand{\corrgm}{$y_{good}$\xspace}

\newcommand{\fbm}{$\V{fb}$\xspace}

\newcommand{\ours}{\textsc{interscript}\xspace}  

\newcommand{\nle}{\textsc{NLEdit}\xspace}

\newcommand{\squishlist}{
  \begin{list}{$\bullet$}
    { \setlength{\itemsep}{0pt}      \setlength{\parsep}{3pt}
      \setlength{\topsep}{3pt}       \setlength{\partopsep}{0pt}
      \setlength{\leftmargin}{1.5em} \setlength{\labelwidth}{1em}
      \setlength{\labelsep}{0.5em} } }
\newcommand{\reallysquishlist}{
  \begin{list}{$\bullet$}
    { \setlength{\itemsep}{0pt}    \setlength{\parsep}{0pt}
      \setlength{\topsep}{0pt}     \setlength{\partopsep}{0pt}
      \setlength{\leftmargin}{0.2em} \setlength{\labelwidth}{0.2em}
      \setlength{\labelsep}{0.2em} } }

 \newcommand{\squishend}{
     \end{list} 
 }
 
\title{\ours: A dataset for interactive learning of scripts through error feedback}
\author{Niket Tandon, Aman Madaan$^\dagger$,  Peter Clark,\\ \textbf{Keisuke Sakaguchi}, Yiming Yang$^\dagger$,\\ 
  Allen Institute for Artificial Intelligence, Seattle, WA, USA \\ 
  $^\dagger$ Language Technologies Institute, Carnegie Mellon University, Pittsburgh, PA, USA \\
  \texttt{\{nikett,peterc,keisukes\}@allenai.org} \\ \texttt{\{amadaan,yiming\}@cs.cmu.edu} }

\begin{document}
\maketitle

\begin{abstract}

How can an end-user provide feedback if a deployed structured prediction model generates inconsistent output,  ignoring the structural complexity of human language? This is an emerging topic with recent progress in synthetic or constrained settings, and the next big leap would require testing and tuning models in real-world settings. We present a new dataset, \ours, containing user feedback on a deployed model that generates complex everyday tasks. \ours\footnote{Code \& data:  \url{https://github.com/allenai/interscript}} contains 8,466 data points-- the input is a possibly erroneous script and a user feedback and the output is a modified script. 
We posit two use-cases of \ours that might significantly advance the state-of-the-art in interactive learning.


\end{abstract}

\section{Introduction}

While language models have achieved remarkable performance on several reasoning tasks \cite{Wang2018GLUE, talmor-etal-2019-commonsenseqa}, they are still prone to mistakes \citep{BenderKoller2020FormAndMeaning}.
This is especially true in structured prediction settings because models can ignore the structural complexity of human
language, and rely on simplistic and error-prone greedy search procedures \cite{martins2020deepspin}. For example, this
leads to critical mistakes in (i) machine translation (such as words being dropped or named entities mistranslated) or
(ii) script generation by a digital assistant to accomplish goals (such as an implausible order of visiting a place and
then driving a car to get there). 

\begin{figure}[!t] \centering \includegraphics[width=\columnwidth]{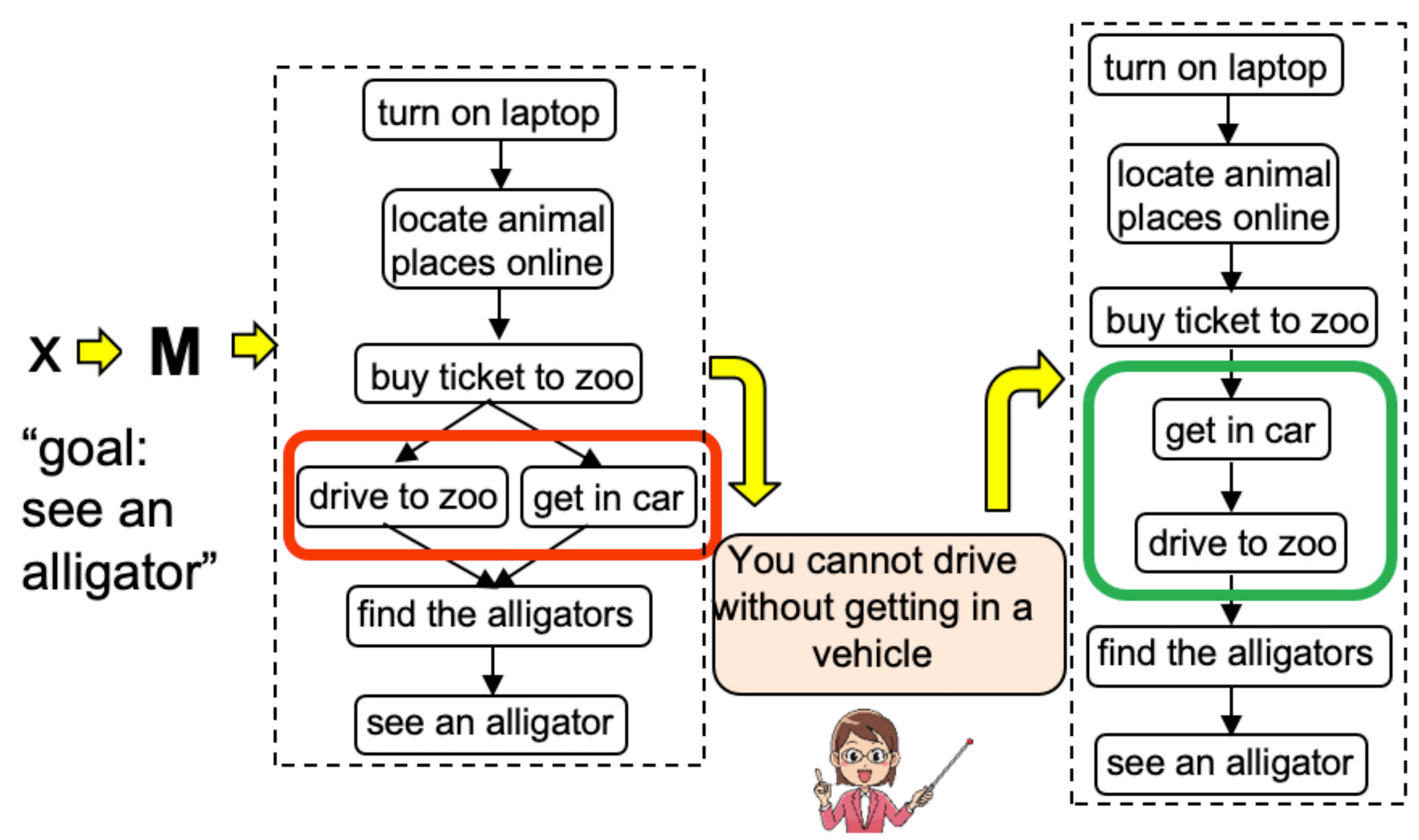} \caption{For an input
goal (see an alligator), the T5-XXL model presented in \cite{proScript} produces an incorrect script, where the order of
the edges is not correct. A human provides a feedback in fluent language. Our dataset, \ours,  contains more than 8K
examples of this kind with the objective of advancing interactive learning.} \label{fig:running_example} \end{figure}


Recent work \cite{proScript} underscores the gap between the syntax and semantic correctness of machine-generated output in the context of automatic script generation. \citet{proScript} report that their models generate scripts that look ostensibly valid.
We conduct an initial study of their generated scripts~(see \S{\ref{sec:subsec:study}}), and find that at least 20\% of the generated scripts contain commonsense mistakes that an average user would be able to point out. This study indicates that such situations can greatly benefit from human assistance in the form of a \textit{feedback} on the error. Figure~\ref{fig:running_example} shows an example of such an erroneous script and the corresponding feedback, where a model-generated script contains an error: it states that the steps of ``driving'' and ``get in the car'' can be applied in any order. The user provides general feedback ``Get in a car before driving''. In a structured setting like script generation, feedback must be localized (e.g., in script generation, this means pointing out the erroneous edge or node in whose context the feedback is provided). 

Teaching a machine via human instructions has long been a goal of AI. Datasets have been crucial in recent AI progress \citep{Wang2018GLUE} and a recent line of work investigates mitigating this problem by curating datasets. Notable datasets to advance interactive learning are RuleTaker \citep{clark2020transformersRuletaker}, TeachYourAI \citep{Talmor2020TeachingPM} and other dialog datasets such as \citet{Padmakumar2021TEAChAmazon, Campos2019ConversationalFAQ}.



Such feedback-driven correction has also been investigated, though in a limited setting, in structured tasks. \citet{percy-2016-learning-language-games-interaction-shouldrn-blockworld} and \citet{Mehta2019interactionRobotUsingAdvice} allow correction to block trajectories in a simulated Block worlds setting. Splash dataset ~\citep{elgohary2021nl,elgohary-etal-2020-splash-dataset} based NLEdit system \citep{elgohary2021nl} allows users to correct semantic parsing errors on database queries using natural language. While encouraging, they are limited in their general applicability due to the synthetic nature of the dataset or task or due to example-specific feedback which does not generalize ~(feedback provided to a query for a database schema might not apply to other queries). Thus, there is still a lack of real-world datasets that can be used to develop systems with feedback-driven correction.

We fill this gap by introducing \ours, a crowdsourced dataset of 8,466 \sample examples, where \badgm is an erroneous script, \fbm the feedback, and \corrgm is the corrected script.
The dataset comprises three types of feedback \fbm, 
(i) explicit (directly mentions the error or localizes the error), 
(ii) implicit (the feedback highlights a general principle which is easy to mentally apply to the given error), 
(iii) distractor feedback (the feedback should not have any effect). Existing datasets provide either explicit or implicit feedback in a synthetic or controlled setting, while \ours is designed to be natural, open, and comprehensive across feedback types. 
We posit two use-cases of \ours on interactive graph correction and continuous learning that might significantly advance the state-of-the-art in interactive learning.

We believe this is a first step towards the larger goal of developing interactive machine learning systems for the widespread structured prediction tasks. 

\section{Related work}

\paragraph{Interactive Learning:} Interactive learning involves a human in the loop, as opposed to learning from datasets collected offline.
Relevant approaches in NLP range from active learning \cite{Raghavan2006ActiveLW,wu2019active} to dialogue systems that adapt to user utterances, spanning diverse domains \cite{Holzinger2016InteractiveMLHealth}. 
There are various modes of interaction (through labels \cite{Raghavan2006ActiveLW, Fails2003InteractiveML}, utterance \cite{Radlinski2019DialogToSeqOfQAPairs}, imitation \cite{Brantley2020ActiveImitationlearning}, and language \cite{elgohary2020speak}). Our work uses language as the mode of interaction. 

\paragraph{Language-based interactions:} Natural language interaction allows for expressive human feedback to correct a model. 
In language-based interactions, controlled settings \cite{Mehta2019interactionRobotUsingAdvice, percy-2016-learning-language-games-interaction-shouldrn-blockworld} give a better handle and are easy to evaluate. However, they do not generalize to real-world settings -- human feedback is rich, and it is not desirable to be restricted to a vocabulary.
Finally, the model being taught is treated either as (i) a black box (as in machine teaching \cite{Dasgupta2019MachineTeaching,Talmor2020TeachingPM}) or (ii) the beliefs of the model are in some form exposed to feedback (as in interactive semantic parsing \cite{elgohary2021nl}). These systems are typically enabled by an underlying dataset. This paper is uniquely positioned because we present the first dataset, which has real user interaction through language in a real-world, open domain, structured prediction setting. 

\paragraph{Interactive Semantic Parsing:} The common theme in prior approaches is based on interactive semantic parsing (such as \citet{elgohary2021nl,percy-2016-learning-language-games-interaction-shouldrn-blockworld}). 
User feedback is mapped into structure edit commands, which can then be executed on the incorrect structures to fix it. For example, \cite{elgohary2021nl} presented \nle to fix SQL queries using human feedback such as:  \texttt{replace course id with program id.}. 
However, the feedback is syntactic with a certain task-specific formal structure, e.g., \nle is known to struggle with natural feedback that does not describe an edit directly \cite{elgohary2021nl}. 
Typically, rather than highlighting a problem or error, these data contain an answer to fix the error. This ``answer containing'' feedback is then parsed using semantic parsing techniques into a set of structure edit commands. 

Unlike NL-Edit, we do not make assumptions about the structure of the feedback. Moreover, we assume that the feedback would be non-actionable (pointing out some local or global error without providing a solution to fix the error). This should especially hold with the growing complexity of the structure to give feedback because it is simpler for a human to point to the problem rather than enumerate (in natural language) the edits that might be required. Further, the feedback in our case is exchangeable. In \nle where the feedback is specific to a query and is grounded in a database schema. In contrast, the feedback in our case addresses a general (commonsense) error with the script, and applies to other scripts with similar issues

\section{Inspiration: Script generation}
Complex everyday events, such as visiting a restaurant, can typically be described as a sequence of distinct sub-events and actions.
Scripts~\citep{Schank1975ScriptsPA} are an efficient way of capturing information about the sub-events and sequences that form a complex event.
A prominent example of scripts in a modern context is their use in digital assistants like Siri, which organize common tasks such as setting up a reminder in script-like workflows~\citep{learning-decompose-tasks-2021}.


The eventual goal of this line of work is, given a goal plan, to execute in a real-world or virtual environment such as ALFRED \cite{Shridhar2020ALFRED}. An intermediate step to reach there is to generate the script given a task such as ``bake a cake'', ``fix a bike tire'' or ``fuel a car''.

\subsection{Script generation}
\begin{figure}[!ht]
\centering
{\includegraphics[width=0.8\columnwidth]{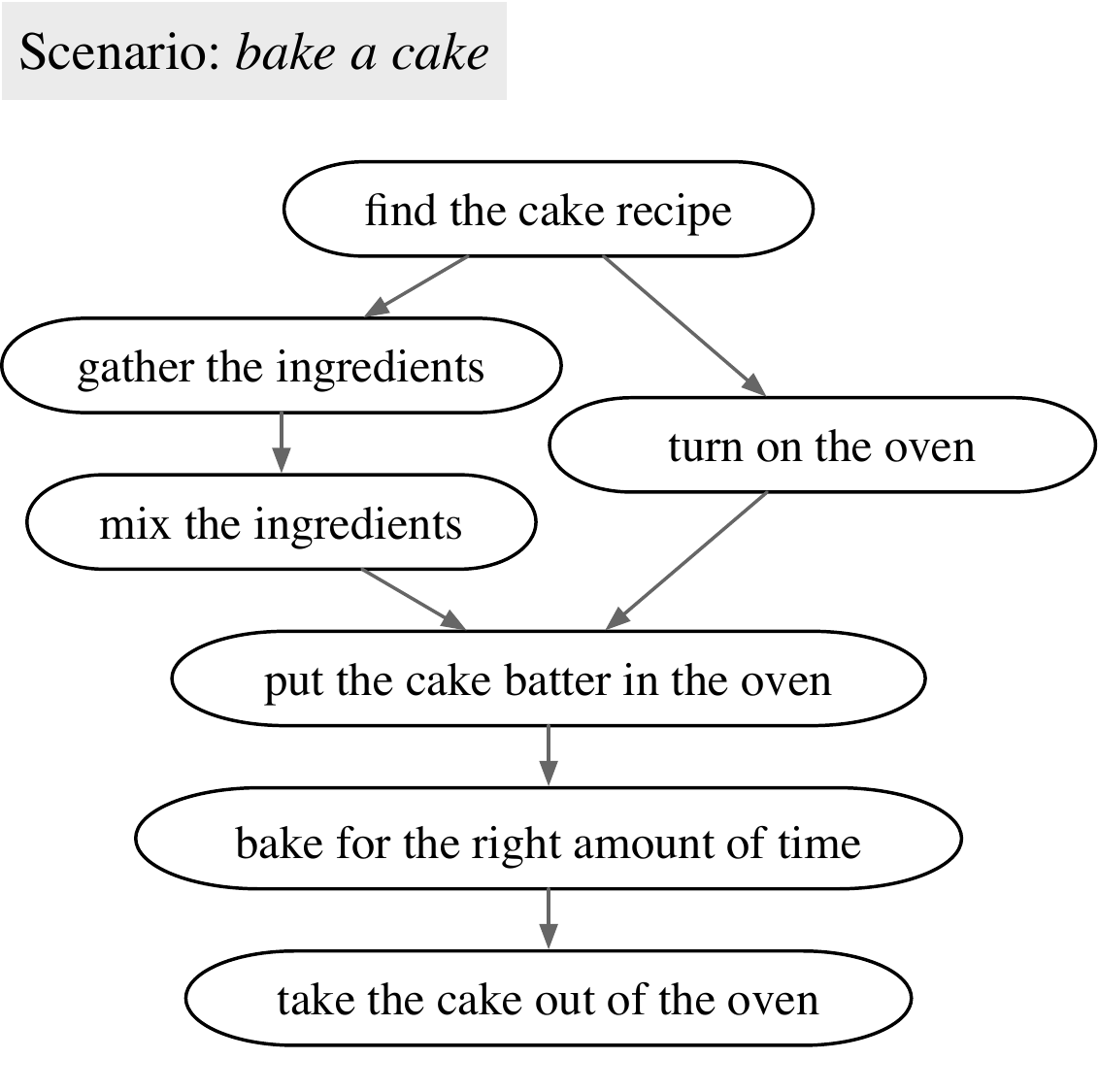}}
\caption{An example of a \textit{script} in \citet{proScript}. In a script generation task, models take the goal as the input and generate a (possibly) partial-order graph, which consists of essential steps and their ordering.}
\label{fig:proscript_orig_example}
\end{figure}

Formally, the script generation task \cite{proScript} takes as input a scenario and generates a script $G(V, E)$, where $V$ is a set of essential events $\{v_1, ... v_i, ... v_{|V|}\}$ and $E$ is a set of temporal ordering constraints between events $\{e_{ij}\}$ which means that the events $v_i$ must precede the event $v_j$ ($v_i \prec v_j$). Partial ordering of events is possible, e.g., you can wear a left sock and a right sock in any temporal order. To solve this task, script generation models are required to \textit{generate} events ($V$) and predict the edges ($E$) jointly. See Figure \ref{fig:proscript_orig_example} for an example.

\proscriptgen  \cite{proScript} is a recently released model that, given a goal, generates $V$ and predicts the edge structure $E$ jointly. It is based on the T5-XXL model (11B parameters) and generates the script as a graph in DOT format. The authors report that the DOT format is always valid at inference time and that $V$ and the graph structure are generally of high quality. They characterize the graph edits required to correct a generated script (such as removing a node, adding a node, changing edge order, etc.). Mechanical Turk workers were able to correct most of the generated scripts within a few edits (typically an edit distance of 5). This makes for an attractive use-case for interactive learning because the generated content from the model (i) is not completely off and (ii) naturally exposes its belief/understanding of the goal through the edge structure, and a user can critique or provide feedback on this belief. 

\subsection{Initial study}
\label{sec:subsec:study}
How do the corrective edits typically look like? On ProScript's test set, we performed inference using the released checkpoint (both GPT-2 and T5-XXL based model). We randomly sampled 30 generated graphs and manually wrote feedback for them (see Table \ref{tab:study-sample-feedback}). 
On average, there were about two mistakes present in the graphs. Often, the error was that the script was using an entity before having it (e.g., write on the paper comes before the node find the paper or reach for the paper). 
Thus, there seems to be a possibility of applying similar feedback to more than one example.
We also found some cases where the script might have to be changed to adapt to special cases. 
For example, for a script \textit{visit Disneyland}, an event \textit{obtain a visa} might be required for some users.
We believe the original ProScript dataset aims to generate widely applicable scripts and grounded in commonsense; rather than cover all possible outcomes. 



\begin{table}[!h]
    \centering
    \begin{tabular}{|p{0.5\columnwidth}|p{0.5\columnwidth}|}
    \hline \textbf{What was the error} &  \textbf{General principle feedback} \\ \hline
    Script was missing the step of not turning off the alarm after waking up & People don't leave their alarms ringing all day.\\ \hline
    Script mentioned coming to the doorway and passing through it & One cannot walk through the doorway without opening the door first. \\ \hline
    Script tells that getting in car and drive in zoo can be done in any order & People must get into a vehicle, before driving to any place. \\ \hline
    Script is looking for a butterfly after placing it & You don't need to look for a butterfly if it's already in a container.\\ \hline
    \end{tabular}
    \caption{Sample feedback on the examples in the study.}
    \label{tab:study-sample-feedback}
\end{table}

\subsection{Conclusions from the initial study}

On the surface, the generated scripts were of good quality.
However, a closer look at the mistakes revealed that most of them could be attributed to the model lacking basic commonsense. 
For example, Figure \ref{fig:running_example} shows a typical mistake the model makes. 
This underscores the gap between the syntax and semantic correctness of machine-generated output in the context of automatic script generation. 
This observation is in-line with other NLP tasks \citep{BenderKoller2020FormAndMeaning} that distinguish the success of recent models on the correctness of form rather than the far-from-over goal of understanding of meaning.

We had two questions: (i) Is there a set of general commonsense principles that the model fails to adhere to. (ii) How can the model incorporate simple feedback and improve? It is costly to retrain large LMs such as the T5-XXL model, and we hoped to make a new component that learns to take feedback from a user. However, we could not find any dataset to solve such an interactive learning problem. Despite recent progress in synthetic or constrained settings, a real-world dataset in a structured prediction setting does not exist. Like with this example, structured prediction models can generate inconsistent output, ignoring the structural complexity of human language. Their decoding is typically based on greedy or sample decoding, and structural incoherence is a possibility. If such a dataset exists with feedback on the generated structure, it would help improve structured outputs-- a widespread problem.
How can an end-user provide feedback if a deployed structured prediction model generates inconsistent output,  ignoring the structural complexity of human language? This is an emerging topic with recent progress in synthetic or constrained settings, and the next big leap would require testing and tuning models in real-world settings. We present a new dataset, \ours, containing user feedback on a deployed model that generates complex everyday tasks. \ours\footnote{Code \& data:  \url{https://anonymous.4open.science/r/interscript}} contains 8,466 data points-- the input is a possibly erroneous script and a user feedback and the output is a modified script. 
We posit two use-cases of \ours that might significantly advance the state-of-the-art in interactive learning.
\section{\ours collection}

\begin{figure}[!ht]
\centering
{\includegraphics[width=0.9\columnwidth]{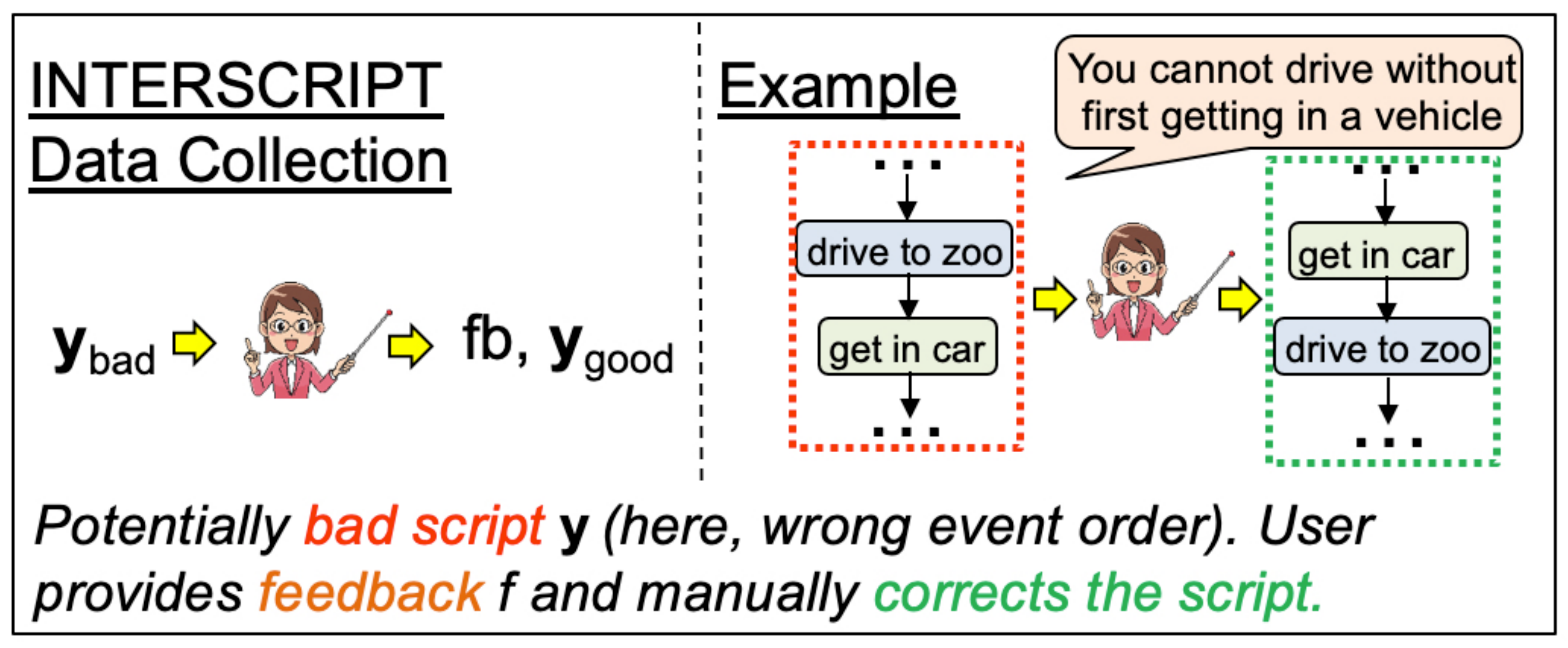}}
\caption{A broad overview of the annotation process. For actual annotation task (including the M-turk task template), see our code repository.
}
\label{fig:mturk-idea}
\end{figure}

We believe that an average user could point out mistakes in the generated scripts, as a majority of the errors in generated scripts are caused by a lack of basic commonsense~(\S{\ref{sec:subsec:study}}).
Consequently, we designed a Mechanical Turk task to provide feedback on mistakes.  A broad overview of the annotation process is shown in Figure \ref{fig:mturk-idea}.

\paragraph{Annotation}
Now we discuss our crowdsourcing setup to collect \ours. To maximize the opportunity to get more feedbacks for a predicted script, we filtered a subset of the test set in ProScript where the human evaluated graph edit distance was likely to be high (i.e., there were likely to be more errors). The ProScript authors released the graph edit value for the set of test set samples they evaluated. We performed inference using their released \proscriptgen model on those data points with high graph edit distance value ($\ge 8$). With this we collected about 400 (predicted graph, expected gold graph) tuples. The ProScript paper describes that their expected gold graph is also imperfect and might contain about 20\% noise. 
Nevertheless, having the gold reference graph guides and constrains an annotator about the common script for a scenario rather than the wide open space of solving the task using multiple potentially correct scripts. (e.g., one could go to a zoo without driving the car by hiring a taxi and then they won't need to drive or park the car). 
As mentioned in \S{\ref{sec:subsec:study}} our annotation process must focus on scripts that are widely applicable and grounded in commonsense.

The annotators are shown the model-generated and expected gold (reference) scripts, and are required to answer which script is worse and why.
It is possible that the gold script is marked as worse.
However, we later post-process and remove such cases, as our focus is to get errors on the generated scripts and not the manually created scripts.
The annotators must point out an egregious mistake (\eg an event or an edge that does not follow commonsense).
They were asked to ignore grammatical and fluency errors, and focus on critical errors of four types:
\squishlist
\item \textbf{Wrong ordering:} the order in the sequence of steps is not correct (e.g., wearing shoes is described before wearing socks).
\item \textbf{Flexible ordering:} some steps can be done in a flexible order (e.g., you can wear left sock or right sock first). A good script captures such flexibility.
\item \textbf{Missing critical steps:} a bad script might have missed critical steps (e.g., the script can say: ``wait for a plane" followed by ``get off the plane" -- here an obvious step ``get on the plane" is missing) . There is no strict definition for a critical step, so the annotators were instructed to use their commonsense judgment.
\item \textbf{Wrong step:} a bad script might have irrelevant and wrong steps (e.g., the script describing ``go to a party" might describe an irrelevant step such as read a book, open a book, etc.).
\squishend

\begin{figure*}[]
\centering
{\includegraphics[width=1.0\linewidth]{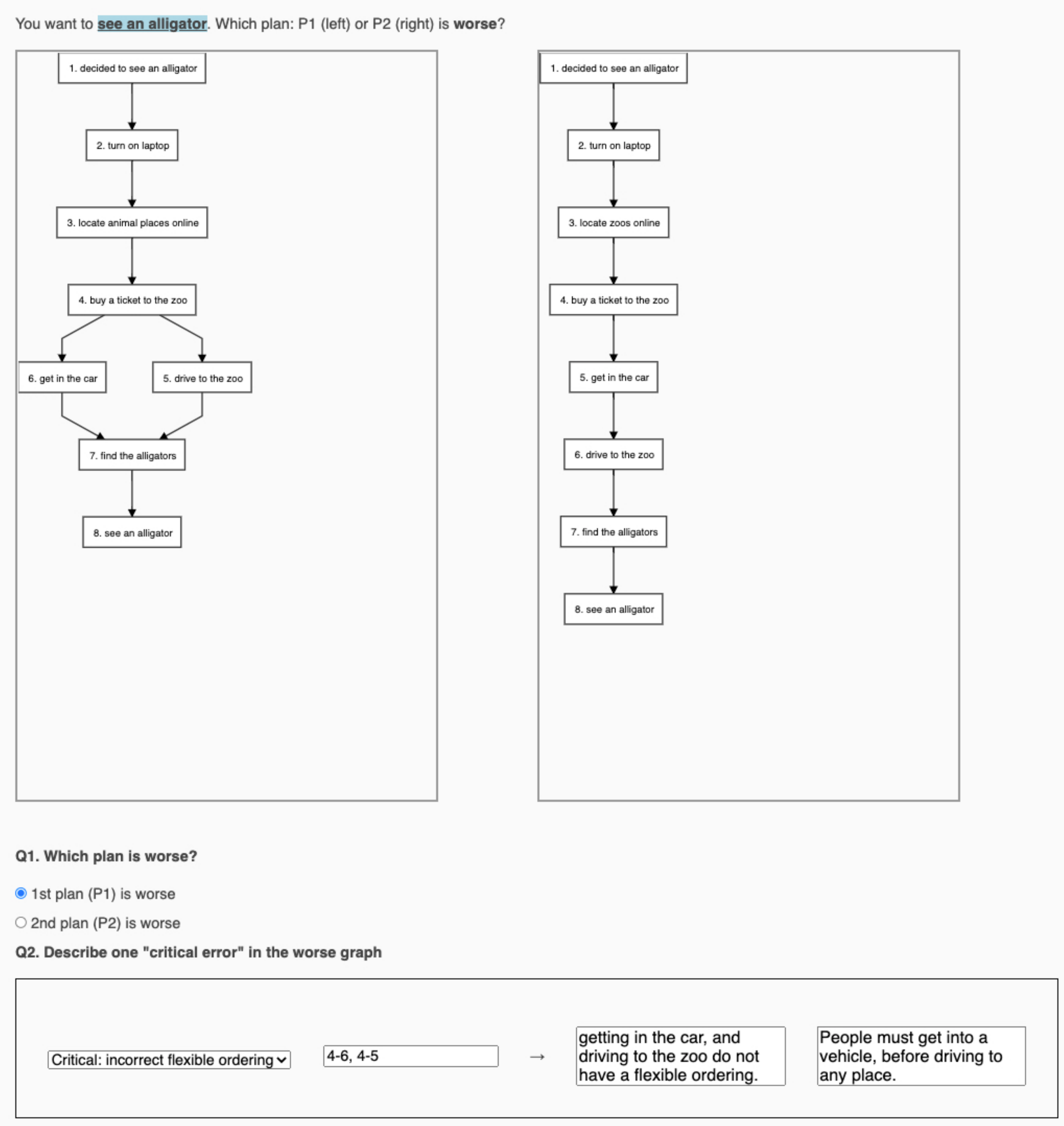}}
\caption{The mechanical turk page for annotation. We show the generated and the expected ProScript gold reference. The annotator must answer which script is worse and why. They must point out an egregious mistake (and not any trivial errors that have minor grammatical errors), and annotate: the error type (missing step, wrong step, wrong order, wrong partial order), localize the error by providing the node or edge id, and give feedback why it is wrong, and finally to gather the general principle behind the feedback they are asked to explain the feedback to a five-year-old.}
\label{fig:mturk-page}
\end{figure*}

For every data point, the annotators were asked to answer the following: 
\begin{enumerate}
    \item Explicit feedback type-1: the error type (missing step, wrong step, wrong order, wrong partial order)
    \item Explicit feedback type-2: localize the error by providing the erroneous node or edge id
    \item Implicit feedback type-1: give feedback in a few words, explaining the error
    \item Implicit feedback type-2: An explanation of the error that would potentially make sense to a five-year-old. This was aimed to explain the feedback to gather the general principle behind the feedback.
\end{enumerate}
Figure \ref{fig:mturk-page} shows a sample of our Mechanical Turk task.

Annotators were required to list only one critical error that they believe was most important.
Each data point is annotated by three annotators, adding some diversity in the errors. 
The annotators were paid \$15 an hour.
Estimated time for completion of one script was 2 minutes. 
We monitored all the jobs via comments, and the annotators did not raise any concerns about the pay or the nature of the task or its difficulty.

We measured the agreement on labels (which graph is worse), and on explicit feedback type-1 and type-2. It was difficult to measure agreement on implicit feedback because it is not easy to perform binary comparison on the generated text without accounting for linguistic variations. On the labels, the Fleiss Kappa agreement was 0.90 (almost perfect agreement) and on explicit feedback the agreement was 0.75 Fleiss Kappa (substantial agreement). This also shows that there is some diversity in what the users perceive as a serious mistake in \badgm. 

Eventually, we compiled these annotations into a dataset of 1,553 tuples of the type \texttt{(explicit feedback type-1, explicit feedback type-2, implicit feedback type-1, implicit feedback type-2)} .

\paragraph{Distractor feedback}
We also add distractor feedbacks, these are cases when the feedback does not apply to the graph. This covers the real-world scenario when a user gives an irrelevant feedback, then an interactive model must be convinced that the feedback is not applicable. To do this, we find the top-k neighbors of the input graph and rest of the graphs by encoding the graphs using Sentence transformers~\citep{reimers-2019-sentence-bert}, and then compute cosine distance. To get irrelevant but lexically related feedback, we attached the feedback of the least similar example (at rank=k and k-1). By random sampling we manually found that k=4 gives reasonable lexical similarity but the feedback are not applicable across the scripts. This subset gives us another approximately 2,026 instances. 


\begin{table}[!h]
    \centering
    \begin{tabular}{|p{0.18\columnwidth}|p{0.1\columnwidth}||p{0.6\columnwidth}|}
    \hline \fbm type &  count & example \\ \hline
explicit \fbm type-1 & 1,553 & Remove node `put the shirt on' \\ \hline
explicit \fbm type-2 & 1,553 & The following step is not right: put the shirt on \\ \hline
implicit \fbm type-1 & 1,553 & It tells you to iron your shirt while it's still on your body. \\ \hline
implicit \fbm type-2 & 1,553 & If you hold a hot iron against the clothes you're currently wearing, you'll get terrible burns. \\ \hline
distractor \fbm      & 2,026 & People do not use two scissors at a time, they simply use one\\  \hline
shared \fbm     & 228   & (illustrative) It tells you to iron your pant while it's still on your body\\ \hline  
total & 8,466 & \url{https://anonymous.4open.science/r/interscript/data.json} \\ \hline
    \end{tabular}
    \caption{\ours dataset statistics.}
    \label{tab:dataset_statistics_by_feedback_type}
\end{table}

\paragraph{Shared feedback}
We often encounter recurring themes or nearly similar situations, and these pairs should share much of their feedback. For example, suppose the goal is to buy grocery and the generated graph is missing an important step like pay for the item. In that case, we can synthesize new examples from it by replacing ``grocery'' with ``toilet paper'' or ``clothes'' in the tuple (\badgm, \fbm, \corrgm).
We manually label a subset of 50 examples by substituting a word in the goal and creating a new tuple. 

We compile these data points into 8,466 samples -- each sample is a tuple (\badgm, \fbm, \corrgm). Table \ref{tab:dataset_statistics_by_feedback_type} breaks down the total number of samples by feedback type. 

We next describe two scenarios where it can potentially advance the field.




\section{Potential Use Cases}

We outline two use cases of \ours that this dataset can potentially enable and advance interactive learning in real-world use cases.

\subsubsection{Use case 1: Learning to apply feedback}
\begin{figure}[!h]
    \centering
    \includegraphics[width=\columnwidth]{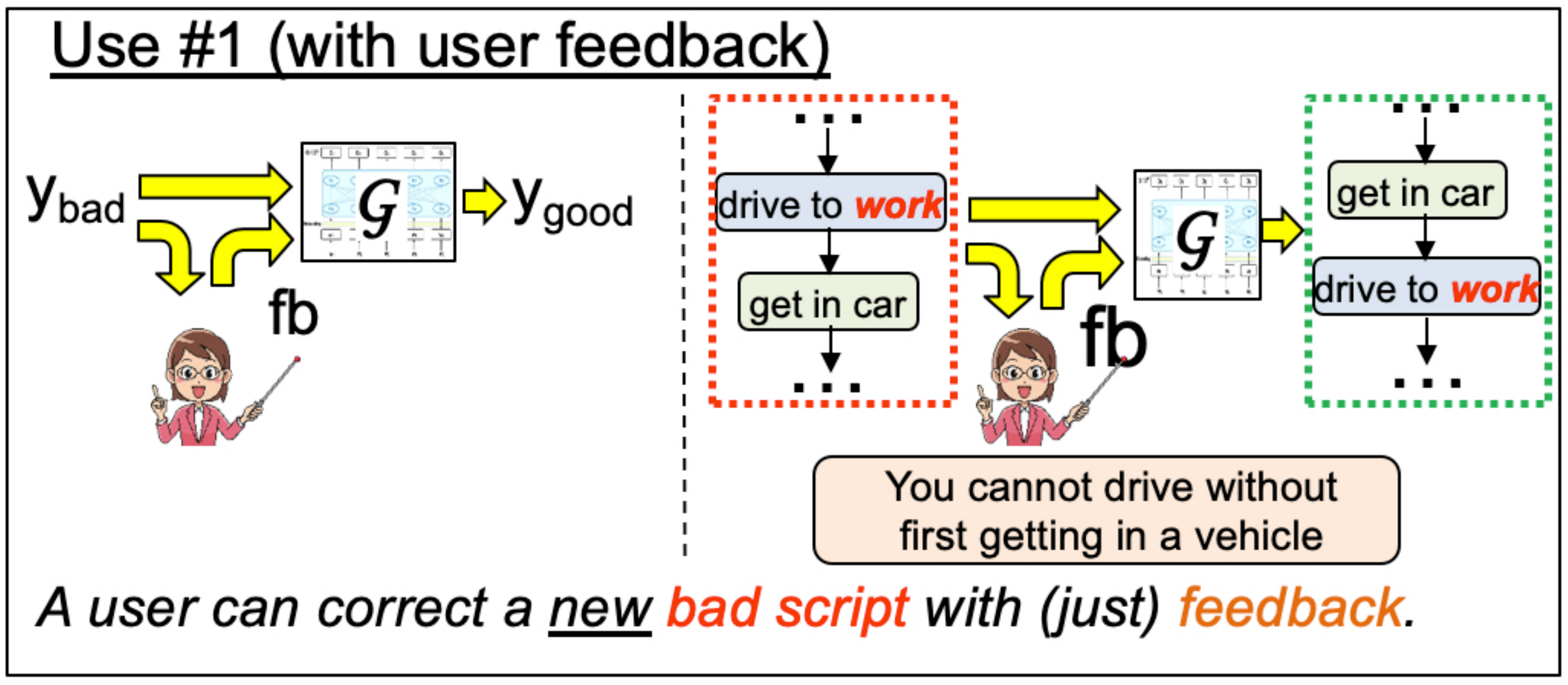}
    \caption{Use case 1: Learning to apply feedback. }
    \label{fig:usecase1}
\end{figure}

The first use case is in training a model that learns to react to user feedback. 
The model can either apply the feedback or suggest that the feedback is not applicable. 
In a real-world structured prediction setting, such a system would be able to correct its answers even after deployment. 
There has been recent interest in this line of work, most recently with OpenAI instruct series\footnote{\url{https://beta.openai.com/docs/engines/instruct-series-beta}} which are specialized to follow a user's instructions. See Figure \ref{fig:usecase1} for an illustrative example.

\subsubsection{Use case 2: Maintaining a memory of feedback.}
A system often encounters recurring themes or situations, or nearly similar situations, e.g., people must get into a vehicle before driving to any place.
Many of the feedback in \ours are based on such general principles that allow the model to use feedback on one context in a different but similar context.


For example, consider a model-generated script for the goal \textit{put a bag in the trunk}, where the model does not include an event \textit{open the trunk} in the generated script.
If the model receives a feedback ``cannot keep something if the container is closed'' for this script, then this error situation is analogous to an error of suggesting to keep a pizza in the oven without opening the oven.
If a model can maintain a memory of errors, it would be possible to envision a system that uses past failures in related error contextscontexts.

\begin{figure}[!h]
    \centering
    \includegraphics[width=\columnwidth]{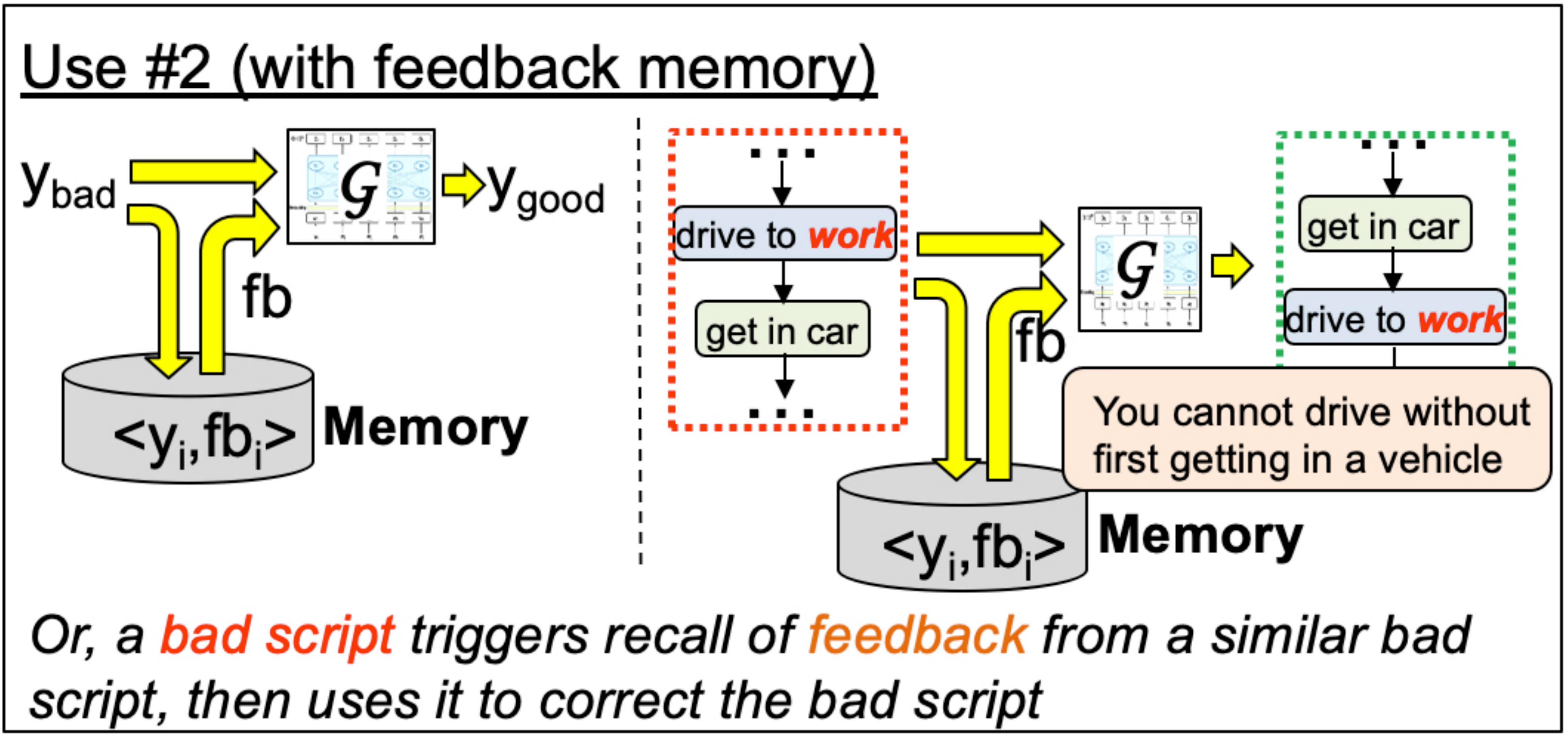}
    \caption{Use case 2: Maintaining a memory of feedback.}
    \label{fig:usecase2}
\end{figure}

Such failure-driven reminding draws inspiration from the ``recursive reminding'' theory in psychology, which suggests that humans remember (in the episodic memory for an event) the context in which they made an error and the received correction. Humans think not only of the error but also of the surrounding context in which it was made and associate it with the correction. In line with this theory, a learner's~(model's) mistakes and the feedback from the user could be in a learner's memory. This will be one instance of a continuous learning model learning with our data (especially the shared feedback subset). Figure \ref{fig:usecase2} presents an illustrative example.

\section{Conclusion} 
We presented \ours, the first real-world dataset for interactive script generation, through error feedback. 
The dataset contains 8,466 data points with a rich hierarchy of feedback types. 
We outlined two use cases that this dataset can potentially enable. These use cases could advance interactive machine learning, especially for very large language models that are difficult to retrain after deployment.

\bibliography{main}

\end{document}